\documentclass[10pt,twocolumn,letterpaper]{article}

\usepackage{cvpr}              

\usepackage{graphicx}
\usepackage{amsmath}
\usepackage{amssymb}
\usepackage{pifont}
\usepackage[table]{xcolor}
\usepackage{multirow}
\usepackage{wrapfig}
\usepackage[pagebackref,breaklinks,colorlinks]{hyperref}
\usepackage{cleveref}
\def\cvprPaperID{XXX} 
\def\confName{CVPR}
\def\confYear{2023}

\begin{document}

\twocolumn[{%
\renewcommand\twocolumn[1][]{#1}%

\vspace{-.2cm}
\title{Segment Anything Is Not Always Perfect: \\ An Investigation of SAM on Different Real-world Applications}
\vspace{-1.5cm}
\author{ \hspace{-1.2ex} Wei Ji \\ \hspace{-1.2ex} University  of Alberta \\ \hspace{-1.2ex} \small{\texttt{wji3@ualberta.ca}}  
		\and 
	    \hspace{-3ex} Jingjing Li \\ \hspace{-3ex} University  of Alberta \\ \hspace{-3ex} \small{\texttt{jingjin1@ualberta.ca}} 
	    \and
	    \hspace{-3ex} Qi Bi \\ \hspace{-3ex} Wuhan University \\ \hspace{-3ex} \small{\texttt{q\_bi@whu.edu.cn}} 
            \and
	   \hspace{-2.9ex} Tingwei Liu \\ \hspace{-2.9ex} Dalian University of Technology \\ \hspace{-2.9ex} \small{\texttt{tingwei@mail.dlut.edu.cn}} 
	    \and
	    Wenbo Li \\ Samsung Research America \\ \small{\texttt{wenbo.li1@samsung.com}} 
	    \and
	    Li Cheng \\ University  of Alberta \\ \small{\texttt{lcheng5@ualberta.ca}} 
}

\maketitle

\begin{center}
    \centering
    \vspace{-0.28cm}
    \includegraphics[width=1\linewidth]{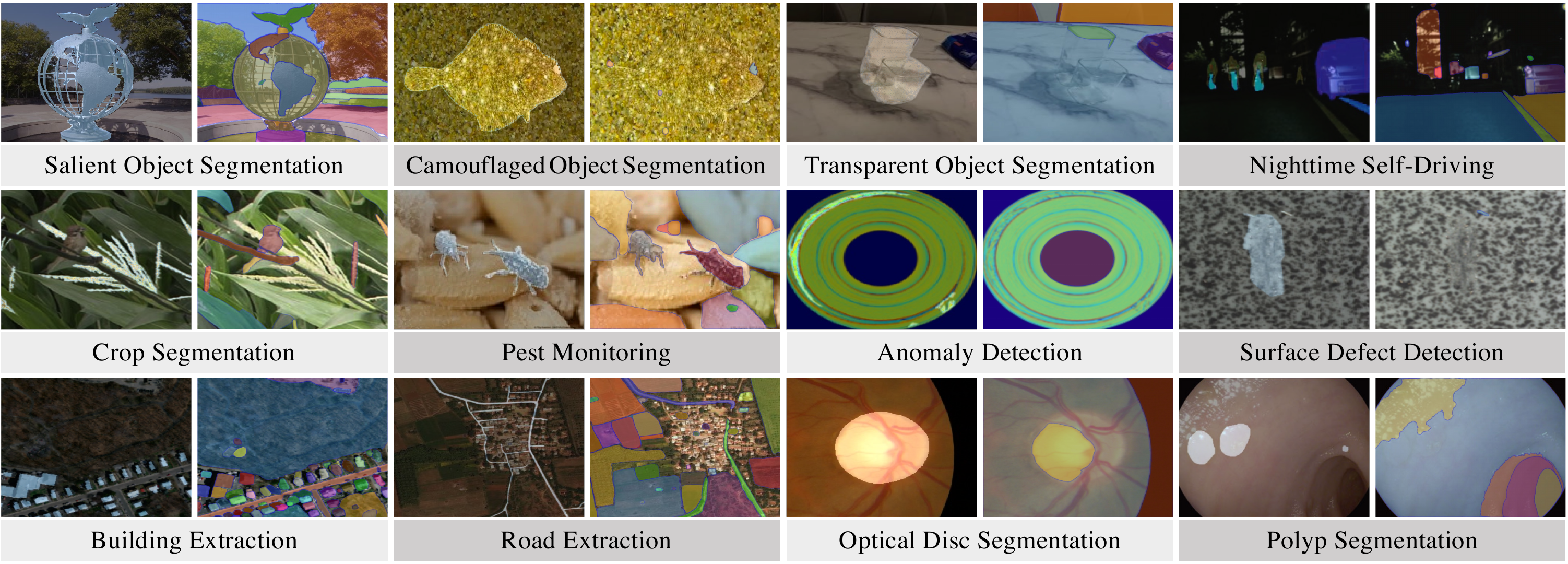}
    \vspace{-6.1mm}
    \captionof{figure}{Results of Segment Anything Model (SAM~\cite{SAM}) on various real-world applications, where we adopt Everything mode to obtain SAM segmentations (\textit{right}). The ground truth is masked with image for reference purpose (\textit{left}). Best view is obtained when zooming in.
    \vspace{.3cm}
    }
    \label{fig:intro}
\end{center}%
}]

\begin{abstract}


Recently, Meta AI Research approaches a general, promptable Segment Anything Model (SAM) pre-trained on an unprecedentedly large segmentation dataset (SA-1B). Without a doubt, the emergence of SAM will yield significant benefits for a wide array of practical image segmentation applications. In this study, we conduct a series of intriguing investigations into the performance of SAM across various applications, particularly in the fields of natural images, agriculture, manufacturing, remote sensing, and healthcare. We analyze and discuss the benefits and limitations of SAM, while also presenting an outlook on its future development in segmentation tasks. By doing so, we aim to give a comprehensive understanding of SAM's practical applications. This work is expected to provide insights that facilitate future research activities toward generic segmentation. Source code is publicly available at \href{https://github.com/LiuTingWed/SAM-Not-Perfect}{here}.

\end{abstract}

\vspace{-0.3cm}
\section{Introduction}
\label{sec:intro}


Growing interest has been observed in foundational models in recent years, which can be attributed to their sufficient pre-training on web-scale datasets and their superior ability to generalize to various downstream tasks.
In natural language processing (NLP), the foundation model \textit{Generative Pre-trained Transformer} (GPT~\cite{GPT}) from OpenAI has demonstrated great generalization ability on many language generative tasks.
Not long after, ChatGPT, empowered by the GPT foundation model, has become a great commercial success, owing to its real-time and reasonable language generation and user interaction. 
Going back to the vision realm, the exploration of foundation models is still in its infancy. The pioneering work of CLIP~\cite{CLIP} effectively combines image-text modalities, enabling zero-shot generalization to novel visual concepts. However, its generalization ability for vision tasks remains unsatisfactory~\cite{SAM} due to the scarcity of abundant training data, unlike in NLP.

More recently, Meta AI Research released a promptable Segment Anything Model (SAM~\cite{SAM}). By incorporating a single user interface as prompt, SAM is capable of segmenting any object in any image or any video without additional training, which is often referred to as zero-shot transfer in the vision community.
As suggested by \cite{SAM}, SAM's capabilities are driven by a vision foundational model that has been trained on a massive SA-1B dataset containing more than 11 million images and one billion masks. 
Meanwhile, the authors have released an impressive online demo to showcase SAM’s capabilities at \url{ https://segment-anything.com}.
SAM is designed to generate a valid segmentation result for any prompt, where prompts can include foreground/background points, a rough box or mask, freeform text, or any other information indicating what to segment in an image.
The latest project offers three prompt modes: click mode, box mode, and everything mode.
Click mode allows users to segment objects with one or more clicks, either including and excluding them from the object.
Box mode allows for object segmentation by roughly drawing a bounding box and using alternative click prompts.
Everything mode automatically identifies and masks all objects in an image.

The emergence of SAM has undoubtedly demonstrated strong generalization across various images and objects, opening up new possibilities and avenues for intelligent image analysis and understanding.
Some practitioners from both industry and academia have gone so far as to assert that `\textit{segmentation has reached its endpoint}' and `\textit{the computer vision community is undergoing a seismic shift}'.
Actually, a dedicated dataset for pre-training is hard to encompass the vast array of unusual real-world scenarios and imaging modalities, particularly for computer vision community with a variety of conditions (\eg, low-light, bird's-eye view, fog, rain), or employing various input modalities (\eg, depth, infrared, event, point cloud, CT, MRI), and with numerous real-world applications.
Thus, it is of great practical interest to investigate how well SAM can infer or generalize under different scenarios and applications. 


This leads us to carry out this study, examining SAM's performance across a diverse range of real-world segmentation applications.
Specifically, we employ SAM\footnote{Visual results are from \url{https://segment-anything.com}. Although we provide effective human prompts in SAM$^{1/2}$ to achieve the best possible results, it should be noted that the optimality of these prompts cannot be guaranteed, as excessive interaction effort is not desirable.} in various practical scenarios~\cite{borji2019salient1,MRNet21,he2019fully,Crop,xu2018road}, including natural image, agriculture, manufacturing, remote sensing and healthcare. Meanwhile, we discuss SAM's benefits and limitations in practice. Based on these studies, we have made the following observations: 
\begin{itemize}
\item

\textit{Excellent generalization on common scenes.} Experiments on various images validate SAM's effectiveness across different prompt modes, demonstrating its ability to generalize well to typical natural image scenarios, especially when target regions distinct prominently from their surroundings. This emphasizes the superiority of the promptable SAM's model design and the strength of its massive and diverse training data source. 

\item 
\textit{Require strong prior knowledge.} During the usage of SAM, we observe that for complex scenes, \eg, crop segmentation and fundus image segmentation, more manual prompts with prior knowledge are required, which could potentially result in a suboptimal user experience. Additionally, we notice that SAM tends to favor selecting the foreground mask. When applying the SAM model to shadow detection task, even with a large number of click prompts, its performance remains poor. This may be due to the strong foreground bias in its pre-training dataset, which hinders its ability to handle certain scenarios effectively.

\item 
\textit{Less effective in low-contrast applications.} 
Segmenting objects with similar surrounding elements is considered challenging scenarios, especially when dealing with transparent or camouflaged objects that are “seamlessly” embedded in their surroundings. Experiments reveal that there is considerable room for exploring and enhancing SAM's robustness in complex scenes with low-contrast elements.

\item 
\textit{Limited understanding of professional data.}
 We apply SAM to real-world medical and industrial scenarios and discover that it produces unsatisfactory results for professional data, particularly when using box mode and everything mode. This reveals SAM's limitations in understanding these practical scenarios. Moreover, even with click mode, both the user and the model are required to possess certain domain-specific knowledge and understanding of the task at hand.

\item 
\textit{Smaller and irregular objects can pose challenges for SAM.} Remote sensing and agriculture present additional challenges, such as irregular buildings and small-sized streets captured from the aerial imaging sensors. These complexities make it challenging for SAM to produce complete segmentation. How to design effective strategies for SAM in such cases is still an open issue.

\end{itemize}

In this study, we examine SAM's performance in various scenarios, and provide some observations and insights toward promoting the development of foundational models in vision realm. While we have tested many tasks, not all downstream applications have been covered. A multitude of fascinating segmentation tasks and scenarios are encouraged to be explored in future research.

\begin{figure*}[t]
\vspace{-0.3cm}
	\centering
	\includegraphics[width=1\linewidth]{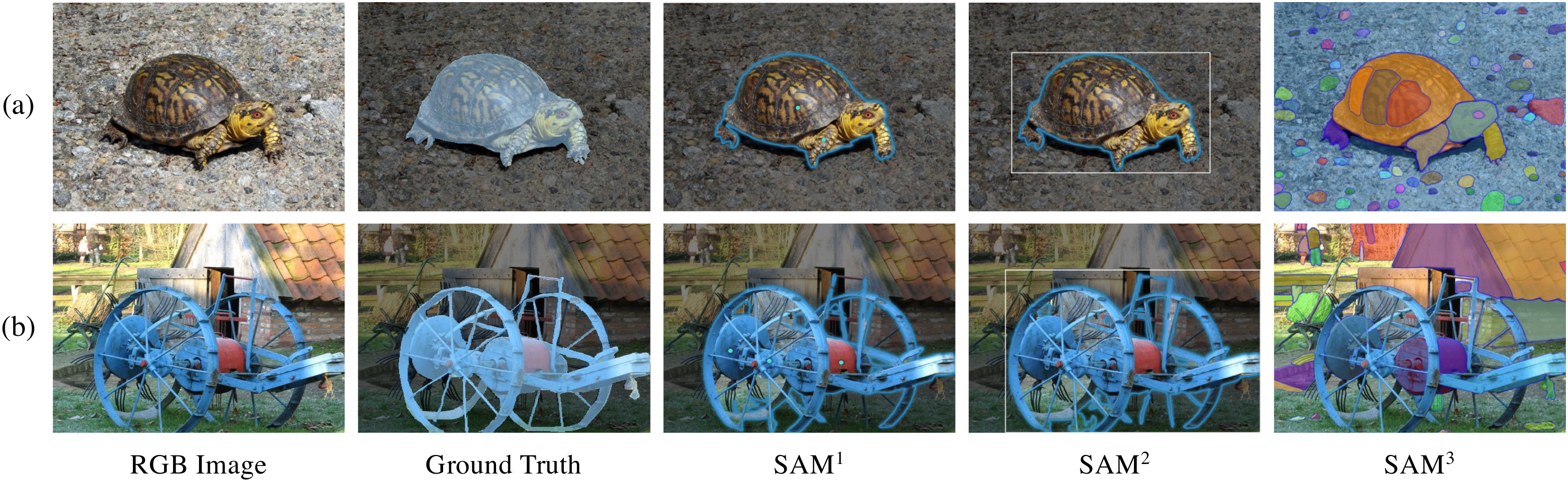}
	\vspace{-0.7cm}
	\caption{Application on \textbf{salient object segmentation}, where SAM$^{1/2/3}$ mean using Click, Box, and Everything modes respectively.}
	\vspace{-0.1cm}
	\label{fig:SOD}
\end{figure*}

\begin{figure*}
	\centering
	\includegraphics[width=1\linewidth]{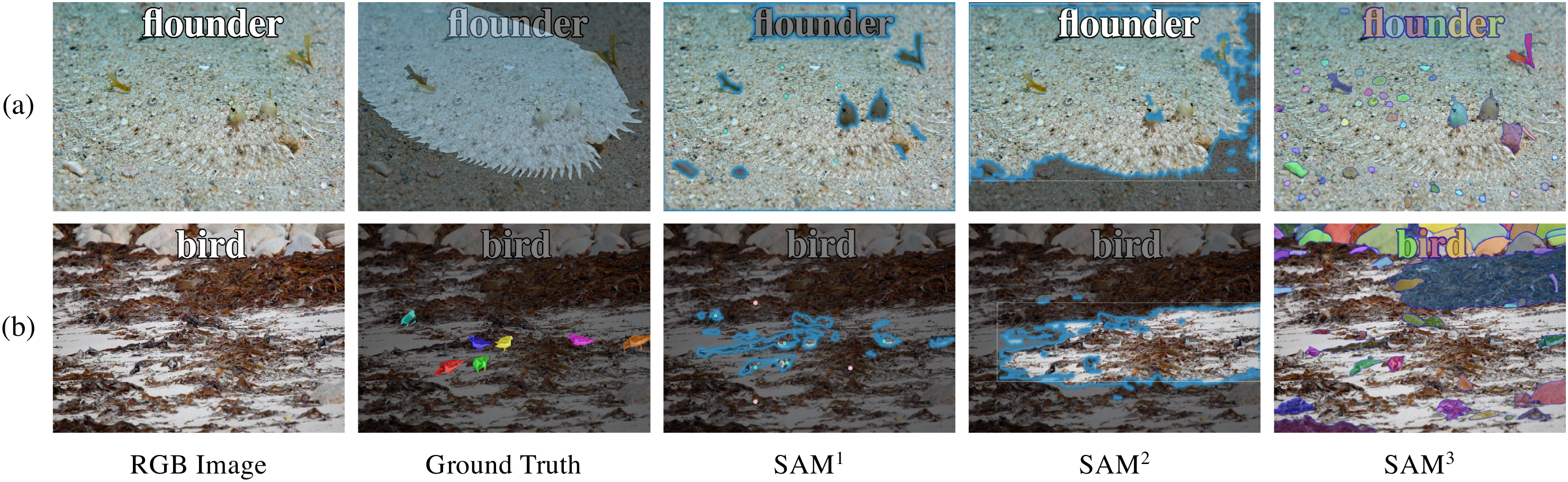}
	\vspace{-0.7cm}
	\caption{Application on \textbf{camouflaged object segmentation}, where SAM$^{1/2/3}$ mean using Click, Box, and Everything modes respectively.}
	\vspace{-0.3cm}
	\label{fig:COD}
\end{figure*}

\begin{figure*}
        \vspace{-0.3cm}
	\centering
	\includegraphics[width=1\linewidth]{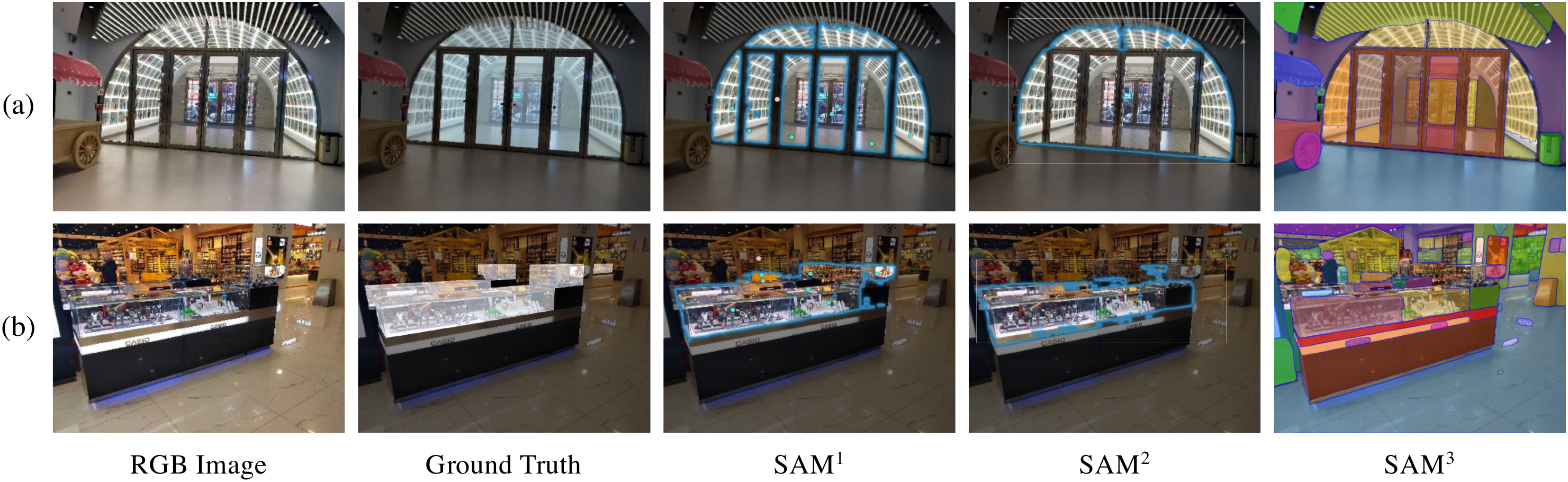}
	\vspace{-0.7cm}
	\caption{Application on \textbf{transparent object segmentation}, where SAM$^{1/2/3}$ mean using Click, Box, and Everything modes respectively.}
	\vspace{-0.1cm}
	\label{fig:Trans}
\end{figure*}

\begin{figure*}
	\centering
	\includegraphics[width=1\linewidth]{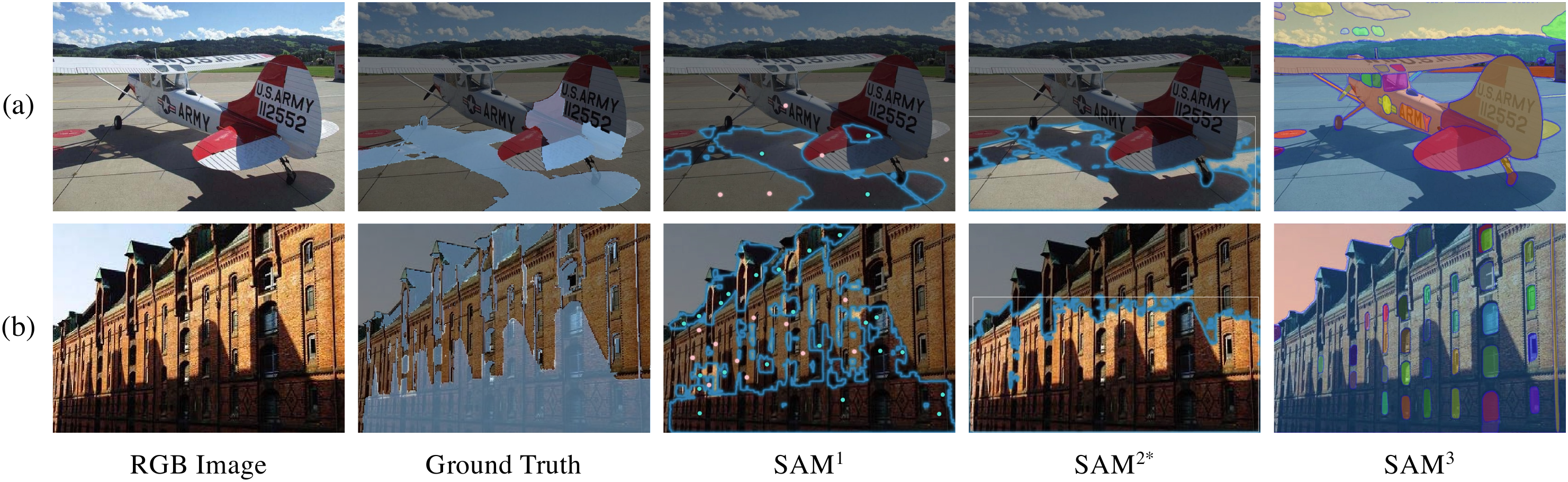}
	\vspace{-0.7cm}
	\caption{Application on \textbf{shadow detection}, where SAM$^{1/2/3}$ mean using Click, Box, and Everything modes respectively. The * indicates the SAM results within a box prompt.}
	\vspace{-0.2cm}
	\label{fig:shadow}
\end{figure*}

\begin{figure*}
        \vspace{-0.5cm}
	\centering
	\includegraphics[width=1\linewidth]{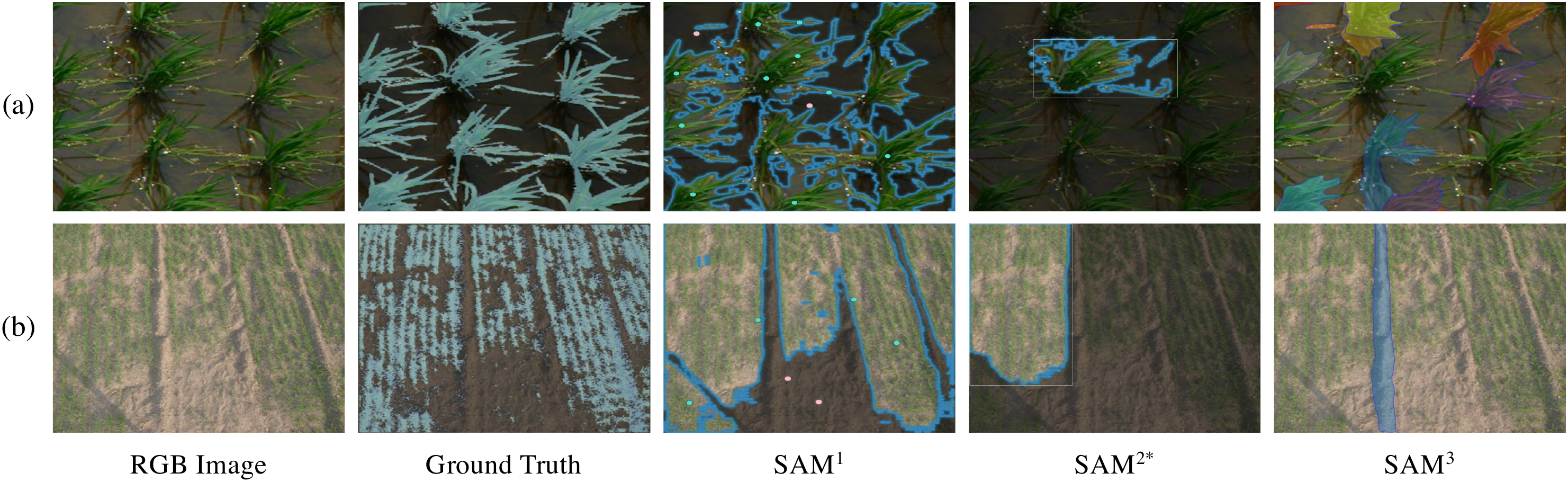}
	\vspace{-0.7cm}
	\caption{Application on \textbf{crop segmentation}, where SAM$^{1/2/3}$ mean using Click, Box, and Everything modes respectively. The * indicates the SAM results within a box prompt.}
	\vspace{-0.1cm}
	\label{fig:Crop}
\end{figure*}

\begin{figure*}
	\centering
	\includegraphics[width=1\linewidth]{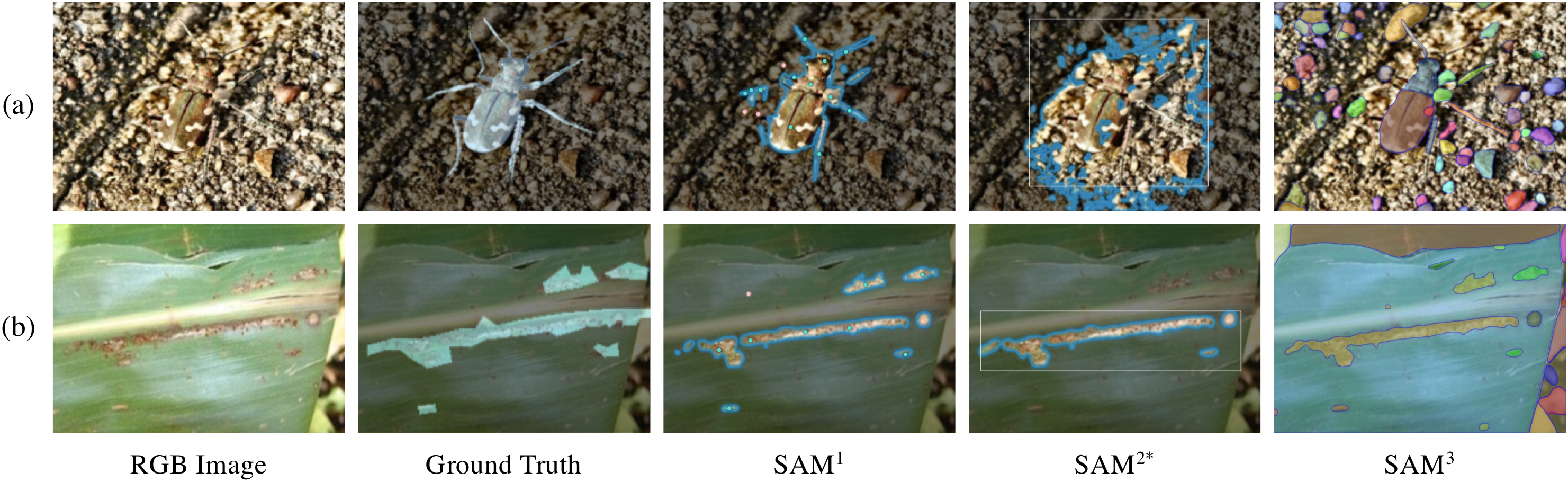}
	\vspace{-0.7cm}
	\caption{Application on \textbf{pest and leaf disease monitoring}, where SAM$^{1/2/3}$ mean using Click, Box, and Everything modes respectively. The * indicates the SAM results within a box prompt.}
	\vspace{-0.5cm}
	\label{fig:Pest}
\end{figure*}

\begin{figure*}
        \vspace{-0.3cm}
	\centering
	\includegraphics[width=1\linewidth]{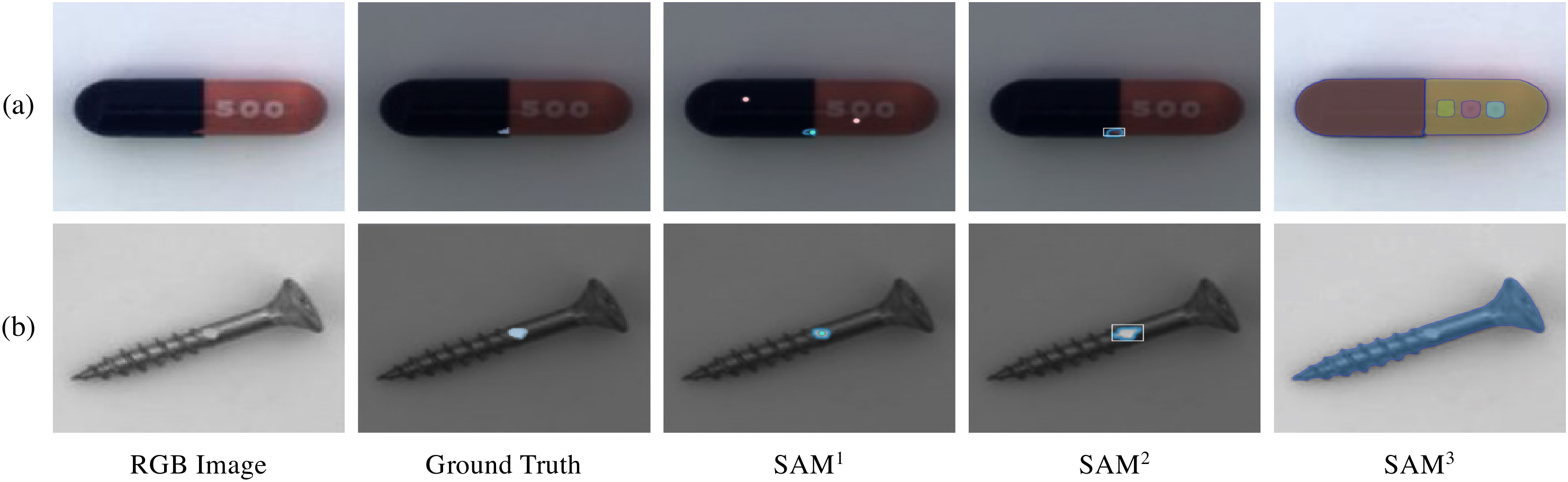}
	\vspace{-0.7cm}
	\caption{Application on \textbf{anomaly detection}, where SAM$^{1/2/3}$ mean using Click, Box, and Everything modes respectively.}
	\vspace{-0.1cm}
	\label{fig:ano}
\end{figure*}

\begin{figure*}
	\centering
	\includegraphics[width=1\linewidth]{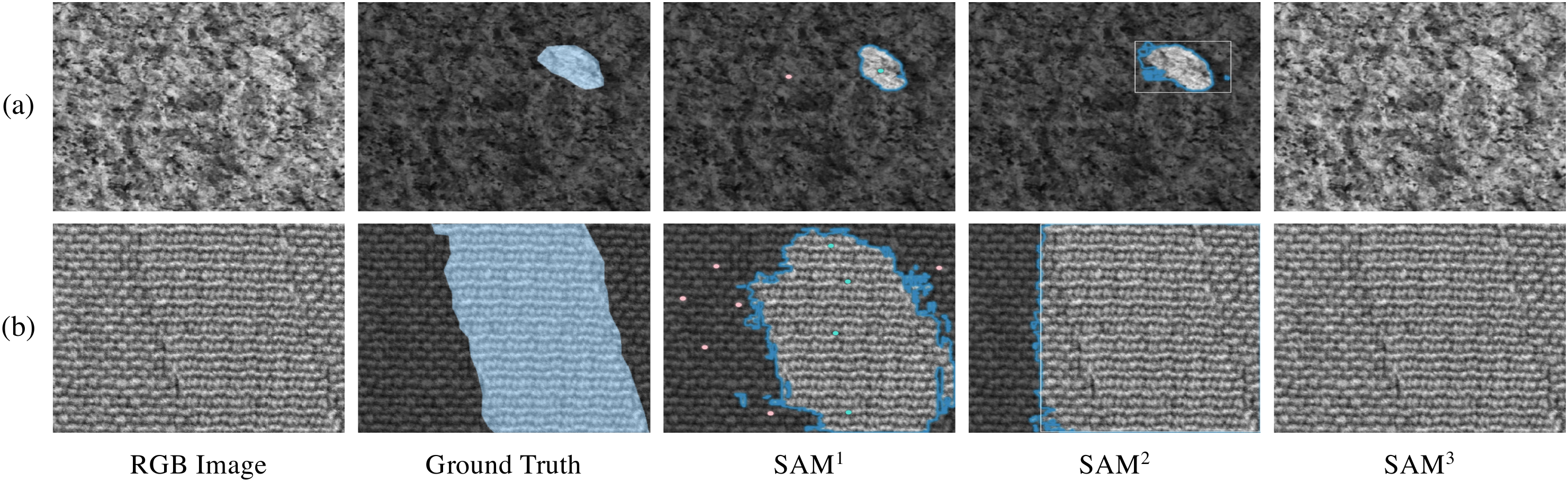}
	\vspace{-0.7cm}
	\caption{Application on \textbf{surface defect detection}, where SAM$^{1/2/3}$ mean using Click, Box, and Everything modes respectively. Here SAM$^{3}$ does not generate any results on these cases.}
	\vspace{-0.4cm}
	\label{fig:defect}
\end{figure*} 

\section{Qualitative Investigation}
\label{sec:inves}
In this section, we present visual results of SAM model on diverse segmentation tasks, and analyze its advantages and limitations from a qualitative perspective.

\subsection{Natural Image Scenes}

\subsubsection{Experiments on Different Subtasks}

\noindent\textbf{Salient Object Segmentation}~\cite{borji2019salient1} aims at extracting the most attention-grabbing objects from still images, differing from semantic segmentation as a class-agnostic task. Here we apply SAM to the popular saliency dataset ReDWeb-S~\cite{WebS}. As shown in Fig.~\ref{fig:SOD} (a), SAM demonstrates a strong ability to identify remarkable objects with precise location. However, when encountering rich-detailed targets, as shown in Fig.~\ref{fig:SOD} (b), SAM is less appealing in capturing fine-grained details.

\noindent\textbf{Camouflaged Object Segmentation}~\cite{COD,tang2023can} involves identifying objects that are `seamlessly' embedded in their surroundings, making it a highly challenging task compared to traditional object segmentation. To verify the effectiveness of SAM, we adopt the COD10K~\cite{COD} test set. As seen in Fig.~\ref{fig:COD}, SAM often fails to detect entire objects in challenging scenarios, such as those with similar foreground and background, or multiple objects in cluttered environments. Even with increased human interactions (\textit{i.e.}, SAM$^1$), many false positives still appear in SAM's results.

\noindent\textbf{Transparent Object Segmentation}~\cite{Trans} is the process of segmenting transparent objects, such as glass, plastic, window, bottle from an image. These objects often exhibit complex shapes and can refract or reflect light in unpredictable ways. Here we apply SAM to the Trans10K-v2 dataset~\cite{Trans} to assess its performance. As shown in Fig.~\ref{fig:Trans}, SAM shows a strong capability to detect the location of target objects, but there is still room for improvement in capturing finer details.

\noindent\textbf{Shadow Detection}~\cite{shadow} is a fundamental problem in which analyzing shadows in a scene can help estimate light conditions and scene geometry. This task is particularly challenging for generic segmentation methods, as the shadows of interest are often difficult to be distinguished from surrounding backgrounds. As shown in Fig.~\ref{fig:shadow}, it is not surprising that SAM fails to detect shadows in the datasets~\cite{shadow,shadow2}.



\vspace{-.3cm}
\subsubsection{Section Recap}
\vspace{-.2cm}
Through assessing on several popular natural image segmentation tasks, we observe that SAM has powerful ability at finding the location of objects, which is particularly impressive for everyday applications such as object detection and online meeting background subtraction. However, in situations involving noisy backgrounds, similar foreground and background, or scenes requiring more detailed segmentation, SAM leaves a significant opportunity for further exploration and improvement.

\subsection{Agriculture}
\subsubsection{Experiments on Different Subtasks}

\noindent\textbf{Crop Segmentation}~\cite{Crop}, a crucial application of image segmentation, is fundamental to agricultural information automation. It serves various purposes, such as predicting crop growth stages, estimating density, identifying cover crops, and monitoring crop biomass. In this study, SAM is tested on dataset~\cite{CropData}, with results displayed in Fig.~\ref{fig:Crop}. In Fig.~\ref{fig:Crop} (a), SAM achieves relatively satisfactory segmentations compared to results in Fig.~\ref{fig:Crop} (b). One possible reason for this discrepancy is that, in general segmentation tasks, crops and land are often treated as background, leading to a scarcity of positive examples that could support SAM in achieving better generalization in agricultural scenes.

\noindent\textbf{Pest and Leaf Disease Monitoring}~\cite{sriwastwa2018detection} is a vital technology in agriculture, contributing to increased food production and quality while reducing plant diseases. Surprisingly, when SAM is applied to the agricultural domain~\cite{Leaf}, it achieves excellent performance, exhibiting robust generalization capabilities as demonstrated in Fig.~\ref{fig:Pest} (b). However, in more challenging scenes like these in Fig.~\ref{fig:Pest} (a), detecting the entire pest body remains difficult for SAM.

\subsubsection{Section Recap}
It is surprising that SAM exhibits appealing performance in the agriculture field, albeit not t hat perfect. We believe that in the future, with sufficient prior knowledge of agriculture, SAM will be able to detect challenging objects more effectively.

\begin{figure*}
	\centering
        \vspace{-0.7cm}
	\includegraphics[width=1\linewidth]{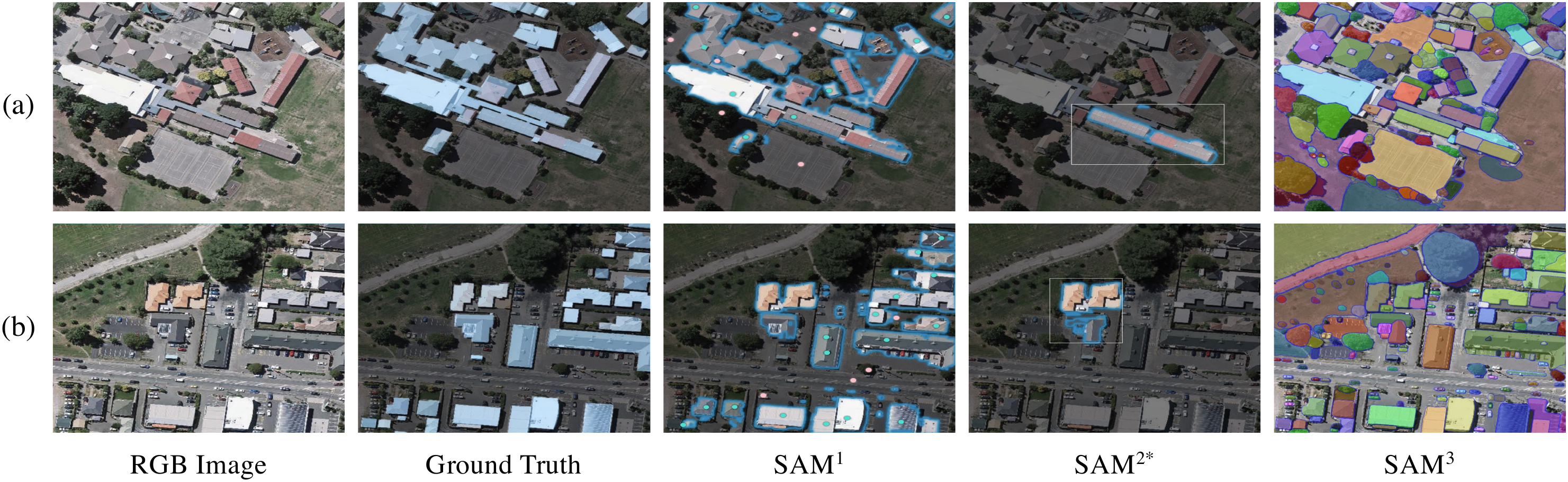}
	\vspace{-0.7cm}
	\caption{Application on \textbf{building extraction}, where SAM$^{1/2/3}$ mean using Click, Box, and Everything modes respectively. The * indicates the SAM results within a box prompt.}
	\vspace{-0.1cm}
	\label{fig:build}
\end{figure*} 

\begin{figure*}
	\centering
	\includegraphics[width=1\linewidth]{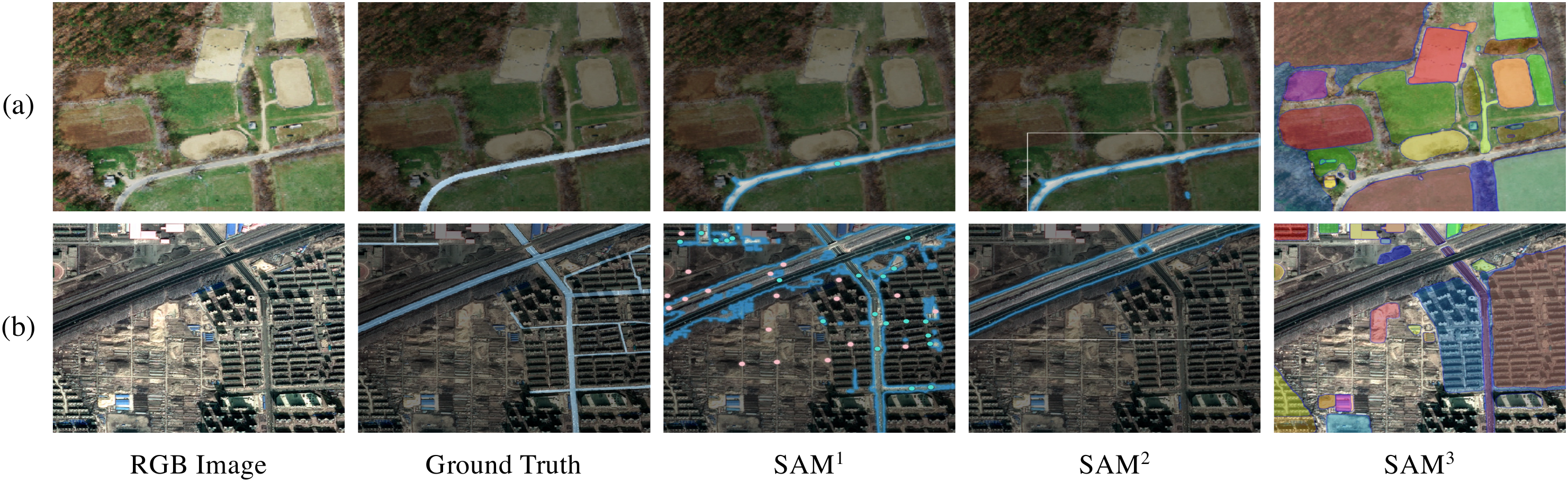}
	\vspace{-0.7cm}
	\caption{Application on \textbf{road extraction}, where SAM$^{1/2/3}$ mean using Click, Box, and Everything modes respectively.}
	\vspace{-0.3cm}
	\label{fig:road}
\end{figure*}

\subsection{Manufacture}
\subsubsection{Experiments on Different Subtasks} 

\noindent\textbf{Anomaly Detection}~\cite{MVTec} is used to detect anomalies from normal samples, which contributes to improved quality control in the industry and increased economic benefits. We test SAM in MVTec AD dataset~\cite{MVTec} as shown in Fig.~\ref{fig:ano}. The results show that it has powerful recognition ability.

\noindent\textbf{Surface Defect Detection}~\cite{he2019fully} aims to identify substandard quality products, such as wood, textiles, and mineral material in industrial manufacturing, to avoid detrimental economic consequences. We further apply SAM to this practical industrial scene. Observations in Fig.~\ref{fig:defect} indicate that, with reasonable human prompts, SAM$^1$ presents acceptable performance, while SAM$^3$ under the everything mode fails to provide a meaningful response. These results suggest that incorporating human experts' prior knowledge is crucial for successful AI model applications in real-world industrial settings.

\vspace{-0.2cm}
\subsubsection{Section Recap}
\vspace{-0.1cm}
Experiments show that SAM performs acceptably in industrial scenarios. In particular, when examining the surface defect detection task, it becomes evident that for practical deployment, a significant amount of expert prior knowledge is indispensable.

\subsection{Remote Sensing}
\subsubsection{Experiments on Different Subtasks}

\noindent\textbf{Building Extraction}~\cite{he2021boundary} from optical remote sensing imagery is one of the fundamental tasks in remote sensing, which plays a key role in many applications, such as urban planning and construction, natural crisis and disaster management \cite{Bi2021LSENet}. As seen in Fig.~\ref{fig:build} (a), SAM is proficient at segmenting regularly shaped objects; however, when dealing with smaller or indiscernible targets as in Fig.~\ref{fig:build} (b), there is still room for improvement.

\noindent\textbf{Road Extraction}~\cite{xu2018road} promotes the development of transportation systems, such as automatic road navigation, unmanned vehicles, and urban planning, which is important in both industry and daily life. We further assess the robustness of SAM using the dataset from \cite{mnih2013machine}. As presented in Fig.~\ref{fig:road}, the objects in the scene exhibit diverse shapes, sizes, and textures, which present challenges for achieving accurate segmentation with SAM.



\vspace{-0.2cm}
\subsubsection{Section Recap}
\vspace{-0.1cm}
Extracting targets from remote sensing images is challenging due to their high variability in shape and size. In our evaluation, we test SAM's performance in extracting essential buildings and roads, and found that it has good scalability for regular objects. However, for segmentation of smaller or less distinguishable objects, it may be necessary to consider task-specific properties and adapt the method accordingly.

\subsection{Healthcare}
\subsubsection{Experiments on Different Subtasks}

\begin{figure*}
	\centering
	\includegraphics[width=1\linewidth]{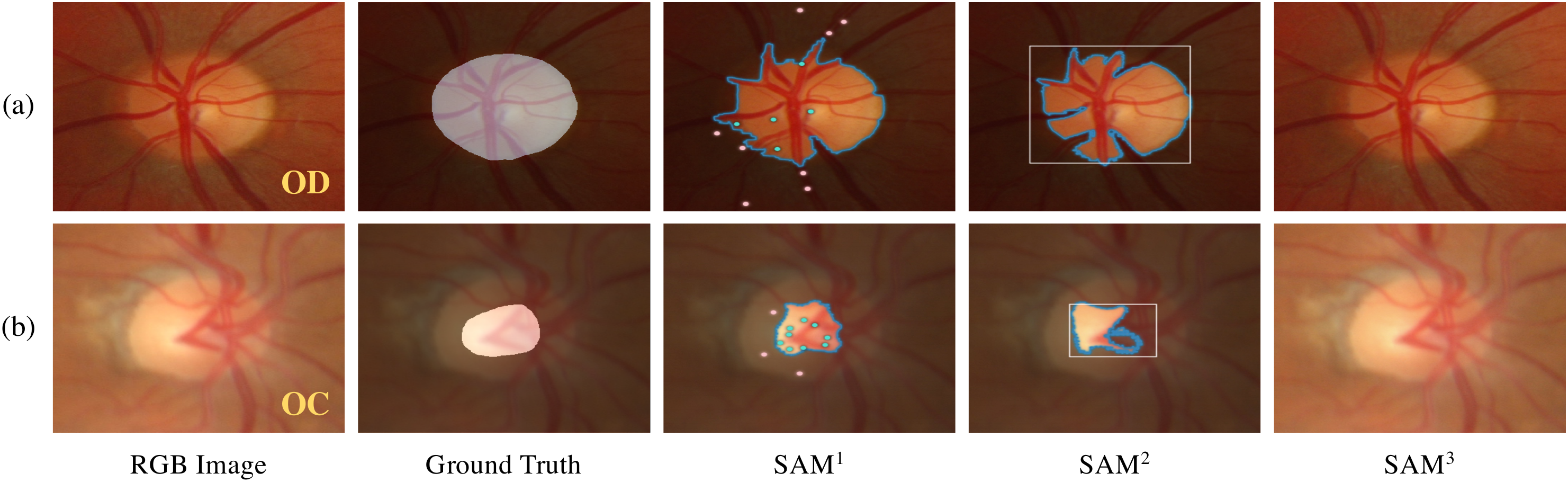}
	\vspace{-0.7cm}
	\caption{Application on \textbf{joint optical disc (OD) and optical cup (OC) segmentation}, where SAM$^{1/2/3}$ mean using Click, Box, and Everything modes respectively. Here SAM$^{3}$ does not generate any results on these cases.}
	\vspace{0.35cm}
	\label{fig:disc}
	\centering
	\includegraphics[width=1\linewidth]{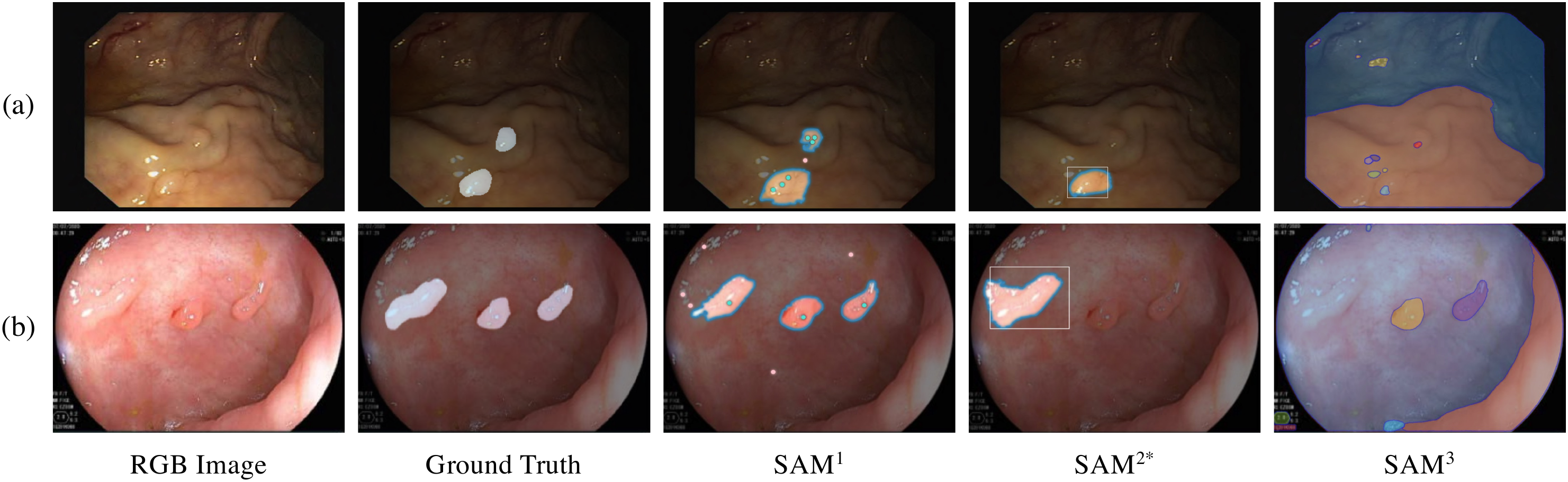}
	\vspace{-0.7cm}
	\caption{Application on \textbf{polyp segmentation}, where SAM$^{1/2/3}$ mean using Click, Box, and Everything modes respectively. The * indicates the SAM results within a box prompt.}
	\vspace{0.35cm}
	\label{fig:polyp}
	\centering
	\includegraphics[width=1\linewidth]{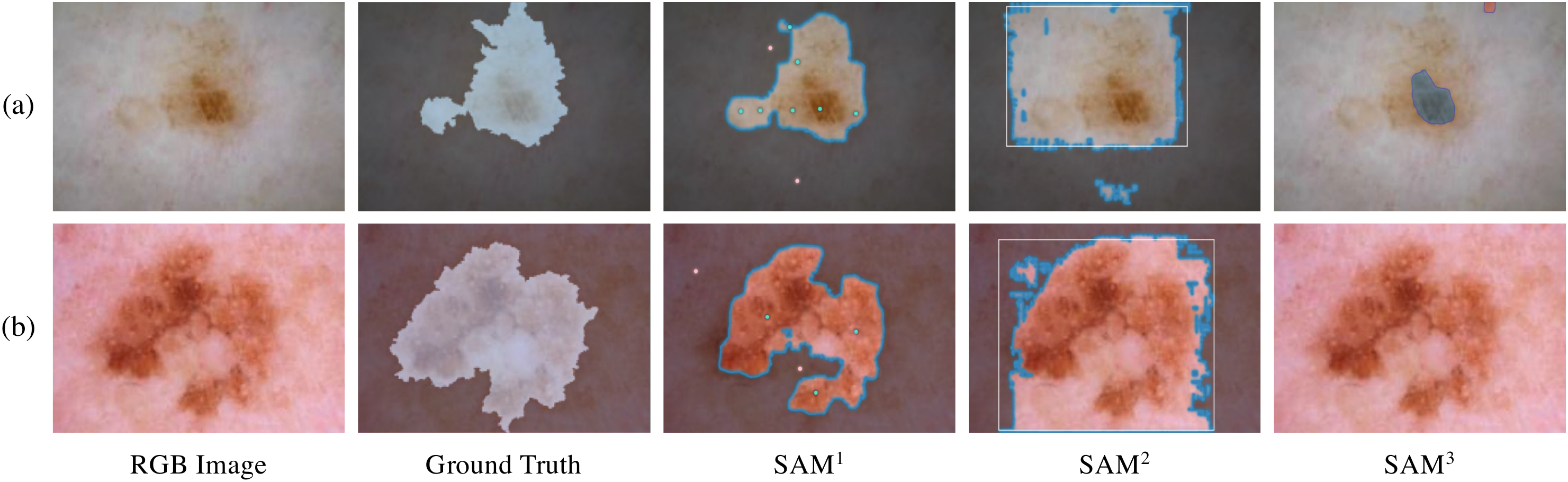}
	\vspace{-0.7cm}
	\caption{Application on \textbf{skin lesion segmentation}, where SAM$^{1/2/3}$ mean using Click, Box, and Everything modes respectively. Here SAM$^{3}$ does not generate any results on the case (b).}
	\vspace{-0.1cm}
	\label{fig:skin}
\end{figure*}

\noindent\textbf{Joint Optical Disc and Cup Segmentation}~\cite{fu2018joint} works towards identifying and separating the optic disc (OD) and the optic cup (OC) in retinal fundus images. The OD is the area where ganglion cell axons exit the eye to form the optic nerve, while the OC is usually restricted to the region inside OD. By jointly segmenting the OD and OC, it can directly compute the cup-to-disc ratio (CDR), an important clinical parameter for the screening of glaucoma. A higher CDR indicates a larger OC and is often associated with a higher risk of glaucoma. In Fig.~\ref{fig:disc}, we evaluate SAM using the popular RIGA benchmark~\cite{RIGA}. The visual results highlight severe shortcomings of the SAM model in medical scenarios. This is understandable, as gold-standard medical image annotations generally require senior ophthalmologists with extensive clinical experience to perform the labeling.

\noindent\textbf{Polyp Segmentation}~\cite{fan2020pranet,pan2022label} is crucial for the early diagnosis and treatment of colorectal cancer. However, polyp segmentation remains a challenging task due to the substantial variation in polyp shape and size. Here we present the results of SAM applied to a poly segmentation benchmark~\cite{an2022blazeneo}, with results displayed in Fig.~\ref{fig:polyp}. It is evident that, with adequate human prompts, SAM$^1$ achieves impressive performance. Nevertheless, for box and automated segmentation modes, SAM$^2$ and SAM$^3$ have difficulty to precisely identify lesion regions. This observation further showcases the ongoing need for significant attention and investigation into the development of dedicated, automated, and effective polyp segmentation algorithms.

\noindent\textbf{Skin Lesion Segmentation}~\cite{codella2018skin} is critical to human life, as the incidence of skin cancer has increased considerably and is seriously threatening human health. This task is particularly challenging due to many factors such as indistinct boundaries, variable contrast, and color differences. Fig.~\ref{fig:skin} presents the results of SAM on~\cite{mendoncca2013ph}. Similar to polyp segmentation, SAM delivers satisfactory outcomes when provided with sufficient human prompts. However, there remains considerable room for improvement in SAM$^2$ and SAM$^3$.



\vspace{-0.2cm}
\subsubsection{Section Recap}
\vspace{-0.1cm}
These results indicate that SAM holds great significance in the medical field. However, it is important to note that SAM$^1$, despite delivering appealing results, often requires substantial human prior knowledge.
This poses a challenge in directly attaining satisfactory results for certain tasks, such as optic cup and disc segmentation, where effective prompts usually require a high level of expertise. 
On the other hand, in box and automatic modes, SAM's performance falls short. This emphasizes the ongoing demand for dedicated SAM models explicitly tailored for tackling medical scenarios.

\section{Quantitative Investigation}
\label{sec:Numerical}

In this section, we present quantitative results of SAM model on diverse segmentation tasks, and compare it with dedicated state-of-the-art models.

\subsection{Datasets and Evaluation Metric}
We thoroughly evaluate SAM on eight representative benchmarks, as follows: 
\\
\textbf{DUTS~\cite{DUTS}} is currently the largest salient object segmentation dataset, consisting of 10,553 training images and 5,019 test images.
\\
\textbf{COME15K-Diff~\cite{COME}} is a challenging salient object segmentation dataset, containing many low-contrast complex scenes. Its test set has 3,000 challenging samples.
\\
\textbf{VT1000~\cite{tu2019rgb}} is a widely-used benchmark for RGB-Thermal salient object segmentation. It covers diverse low-light images, and can be used to assess the robustness of SAM in night scenes. 
\\
\textbf{DIS-TE4~\cite{DIS}} provides 500 highly accurate pixel-level masks for images with complex object shapes. This can evaluate SAM's ability in perceiving boundary details.
\\
\textbf{COD10K~\cite{COD}} is a substantial dataset specifically designed for camouflaged object segmentation, featuring 10,000 images characterized by similar foreground objects and surrounding environments.
\\
\textbf{SBU~\cite{SBU}} provides carefully annotated shadow masks for 700 images, which is widely used in shadow detection task. 
\\
\textbf{CDS2K~\cite{CSU23}} is a comprehensive concealed defect segmentation dataset, compiled from a variety of established industrial defect repositories. It comprises 2,492 samples, consisting of 1,330 positive and 1,162 negative instances.
\\
\textbf{ColonDB~\cite{tajbakhsh2015automated}} is used to verify the generalization capability of SAM in medical field, especially in the challenging polyp lesion segmentation. It consists of 380 images in total.

For quantitative evaluation, we adopt the widely-used mean absolute error ($\mathcal{M}$). A lower $\mathcal{M}$ score indicates better model performance.

\begin{table*}
\vspace{.5cm}
  \centering
    \begin{subtable}{.245\linewidth}
    \centering
    \caption{Results on DUTS~\cite{DUTS}. }
\resizebox{!}{1.6 cm}{    
\begin{tabular}{c|c|c}
    \hline
    {Model} & {Backbone} & $\mathcal{M}$ \\
    \hline
    VST$_{21}$~\cite{VST} & T2T-ViTt-14 & 0.037 \\
    ICONet$_{22}$~\cite{zhuge2022salient} & ResNet50 & 0.037 \\
    EDNet$_{22}$~\cite{wu2022edn} & ResNet50 & \textbf{0.035} \\
    \hline
    \multirow{6}[2]{*}{SAM$_{23}$} & ViT-B & 0.121 \\
          & \small{$\Delta$\textit{diff}}			   & \small{\textcolor[rgb]{ .459,  .443,  .443}{$\downarrow$8.6\%}} \\ \cline{2-3}
          & ViT-L 					   & 0.052 \\
          & \small{$\Delta$\textit{diff}}				  &  \small{\textcolor[rgb]{ .459,  .443,  .443}{$\downarrow$1.7\%}} \\ \cline{2-3}
          & ViT-H 					   & 0.046 \\
          & \small{$\Delta$\textit{diff}}			         & \small{\textcolor[rgb]{ .459,  .443,  .443}{$\downarrow$1.1\%}}   \\
    \hline
    \end{tabular}}
  \end{subtable}
  \begin{subtable}{.245\linewidth}
    \centering
    \caption{Results on COME15K-Diff~\cite{COME}. }
\resizebox{!}{1.6 cm}{    
\begin{tabular}{c|c|c}
    \hline
    {Model} & {Backbone} & $\mathcal{M}$ \\
    \hline
    VST$_{21}$~\cite{VST} & T2T-ViTt-14 & 0.054 \\
    ICONet$_{22}$~\cite{zhuge2022salient} & ResNet50 & 0.055 \\
    EDNet$_{22}$~\cite{wu2022edn} & ResNet50 & \textbf{0.050} \\
    \hline
    \multirow{6}[2]{*}{SAM$_{23}$} & ViT-B & 0.129 \\
          & \small{$\Delta$\textit{diff}}			   & \small{\textcolor[rgb]{ .459,  .443,  .443}{$\downarrow$7.9\%}} \\ \cline{2-3}
          & ViT-L 					   & 0.061 \\
          & \small{$\Delta$\textit{diff}}				  &  \small{\textcolor[rgb]{ .459,  .443,  .443}{$\downarrow$1.1\%}} \\ \cline{2-3}
          & ViT-H 					   & 0.058 \\
          & \small{$\Delta$\textit{diff}}			         & \small{\textcolor[rgb]{ .459,  .443,  .443}{$\downarrow$0.8\%}}   \\
    \hline
    \end{tabular}}
  \end{subtable}
    \begin{subtable}{.242\linewidth}
    \centering
   \caption{Results on VT1000~\cite{tu2019rgb}.}
\resizebox{!}{1.6 cm}{    
\begin{tabular}{c|c|c}
    \hline
    {Model} & {Backbone} & $\mathcal{M}$ \\
    \hline
    VST$_{21}$~\cite{VST} & T2T-ViTt-14 & \textbf{0.024} \\
    ICONet$_{22}$~\cite{zhuge2022salient} & ResNet50 & 0.032 \\
    EDNet$_{22}$~\cite{wu2022edn} & ResNet50 & {0.028} \\
    \hline
    \multirow{6}[2]{*}{SAM$_{23}$} & ViT-B & 0.060 \\
          & \small{$\Delta$\textit{diff}}			   & \small{\textcolor[rgb]{ .459,  .443,  .443}{$\downarrow$3.6\%}} \\ \cline{2-3}
          & ViT-L 					   & 0.035 \\
          & \small{$\Delta$\textit{diff}}				  &  \small{\textcolor[rgb]{ .459,  .443,  .443}{$\downarrow$1.1\%}} \\ \cline{2-3}
          & ViT-H 					   & 0.032 \\
          & \small{$\Delta$\textit{diff}}			         & \small{\textcolor[rgb]{ .459,  .443,  .443}{$\downarrow$0.8\%}}   \\
    \hline
    \end{tabular}}
  \end{subtable}
    \begin{subtable}{.242\linewidth}
    \centering
    \caption{Results on DIS-TE4~\cite{DIS}. }
\resizebox{!}{1.6 cm}{    
\begin{tabular}{c|c|c}
    \hline
    {Model} & {Backbone} & $\mathcal{M}$ \\
    \hline
    Gate$_{20}$~\cite{zhao2020suppress} 	& ResNet50  	& 0.109 \\
    PFNet$_{21}$~\cite{mei2021camouflaged} 		& ResNet50 	& 0.107 \\
    IS-Net$_{22}$~\cite{DIS} 	& U2Net & \textbf{0.072} \\
    \hline
    \multirow{6}[2]{*}{SAM$_{23}$} & ViT-B & 0.179 \\
          & \small{$\Delta$\textit{diff}}			   & \small{\textcolor[rgb]{ .459,  .443,  .443}{$\downarrow$10.7\%}} \\ \cline{2-3}
          & ViT-L 					   & 0.166 \\
          & \small{$\Delta$\textit{diff}}				  &  \small{\textcolor[rgb]{ .459,  .443,  .443}{$\downarrow$9.4\%}} \\ \cline{2-3}
          & ViT-H 					   & 0.166 \\
          & \small{$\Delta$\textit{diff}}			         & \small{\textcolor[rgb]{ .459,  .443,  .443}{$\downarrow$9.4\%}}   \\
    \hline
    \end{tabular}}
  \end{subtable}
  \begin{subtable}{.242\linewidth}
    \centering
    \caption{Results on COD10K~\cite{COD}. }
\resizebox{!}{1.6 cm}{    
\begin{tabular}{c|c|c}
    \hline
    {Model} & {Backbone} & $\mathcal{M}$  \\
    \hline
    SegMaR$_{22}$~\cite{jia2022segment} 	& ResNet50 	& 0.034  \\
    PFNet+$_{23}$~\cite{PFNet_Plus} 	& ResNet50 & 0.037 \\
    ZoomNet$_{22}$~\cite{pang2022zoom} 	&ResNet50 & \textbf{0.029} \\
    \hline
    \multirow{6}[2]{*}{SAM$_{23}$} & ViT-B & 0.108 \\
          & \small{$\Delta$\textit{diff}}			   & \small{\textcolor[rgb]{ .459,  .443,  .443}{$\downarrow$7.9\%}} \\ \cline{2-3}
          & ViT-L 					   & 0.065 \\
          & \small{$\Delta$\textit{diff}}				  &  \small{\textcolor[rgb]{ .459,  .443,  .443}{$\downarrow$3.6\%}} \\ \cline{2-3}
          & ViT-H 					   & 0.054 \\
          & \small{$\Delta$\textit{diff}}			         & \small{\textcolor[rgb]{ .459,  .443,  .443}{$\downarrow$2.5\%}}   \\
    \hline
    \end{tabular}}
  \end{subtable}
    \begin{subtable}{.242\linewidth}
    \centering
   \caption{Results on SBU~\cite{SBU}.}
\resizebox{!}{1.6 cm}{    
\begin{tabular}{c|c|c}
    \hline
    {Model} & {Backbone} & $\mathcal{M}$  \\
    \hline
    DSC$_{18}$~\cite{hu2018direction} & VGG-16  & 0.032 \\
    DSDNet$_{19}$~\cite{zheng2019distraction} & ResNext & {0.036} \\
    MTMT$_{20}$~\cite{chen2020multi} & ResNext & \textbf{0.029} \\
    \hline
    \multirow{6}[2]{*}{SAM$_{23}$} & ViT-B & 0.203 \\
          & \small{$\Delta$\textit{diff}}			   & \small{\textcolor[rgb]{ .459,  .443,  .443}{$\downarrow$17.4\%}} \\ \cline{2-3}
          & ViT-L 					   & 0.187 \\
          & \small{$\Delta$\textit{diff}}				  &  \small{\textcolor[rgb]{ .459,  .443,  .443}{$\downarrow$15.8\%}} \\ \cline{2-3}
          & ViT-H 					   & 0.183 \\
          & \small{$\Delta$\textit{diff}}			         & \small{\textcolor[rgb]{ .459,  .443,  .443}{$\downarrow$15.4\%}}   \\
    \hline
    \end{tabular}}
  \end{subtable}
    \begin{subtable}{.242\linewidth}
    \centering
    \caption{Results on CDS2K~\cite{CSU23}. }
\resizebox{!}{1.6 cm}{    
\begin{tabular}{c|c|c}
    \hline
    {Model} & {Backbone} & $\mathcal{M}$ \\
    \hline
    SINetV2$_{22}$~\cite{fan2021concealed} 	& Res2Net50  	& 0.102 \\
    HitNet$_{23}$~\cite{HiNet} 		& PVTv2-B2 	& 0.118  \\
    DGNet$_{23}$~\cite{DGNet} 	& EffiNet-B4 & \textbf{0.089} \\
    \hline
    \multirow{6}[2]{*}{SAM$_{23}$} & ViT-B & 0.372 \\
          & \small{$\Delta$\textit{diff}}			   & \small{\textcolor[rgb]{ .459,  .443,  .443}{$\downarrow$28.3\%}} \\ \cline{2-3}
          & ViT-L 					   & 0.281 \\
          & \small{$\Delta$\textit{diff}}				  &  \small{\textcolor[rgb]{ .459,  .443,  .443}{$\downarrow$19.2\%}} \\ \cline{2-3}
          & ViT-H 					   & 0.265 \\
          & \small{$\Delta$\textit{diff}}			         & \small{\textcolor[rgb]{ .459,  .443,  .443}{$\downarrow$17.6\%}}   \\
    \hline
    \end{tabular}}
  \end{subtable}
    \begin{subtable}{.242\linewidth}
    \centering
   \caption{Results on ColonDB~\cite{tajbakhsh2015automated}.}
\resizebox{!}{1.6 cm}{    
\begin{tabular}{c|c|c}
    \hline
    {Model} & {Backbone} & $\mathcal{M}$ \\
    \hline
    FAPNet$_{22}$~\cite{zhou2022feature} & Res2Net50  & 0.038 \\
    CFA-Net$_{23}$~\cite{zhou2023cross} & Res2Net50 & {0.039} \\
    HSNet$_{22}$~\cite{zhang2022hsnet} &  PVTv2 & \textbf{0.032} \\
    \hline
    \multirow{6}[2]{*}{SAM$_{23}$} & ViT-B & 0.111 \\
          & \small{$\Delta$\textit{diff}}			   & \small{\textcolor[rgb]{ .459,  .443,  .443} {$\downarrow$7.9\%}} \\ \cline{2-3}
          & ViT-L 					   & 0.054 \\
          & \small{$\Delta$\textit{diff}}				  &  \small{\textcolor[rgb]{ .459,  .443,  .443}{$\downarrow$2.2\%}} \\ \cline{2-3}
          & ViT-H 					   & 0.056 \\
          & \small{$\Delta$\textit{diff}}			         & \small{\textcolor[rgb]{ .459,  .443,  .443}{$\downarrow$2.4\%}}   \\
    \hline
    \end{tabular}}
  \end{subtable}
    \caption{Quantitative results of SAM on applications of (a) salient object segmentation in \textit{common scenes}, (b) salient object segmentation in \textit{low-contrast scenes}, (c) salient object segmentation in \textit{low-light scenes}, (d) salient object segmentation with \textit{highly-accurate details} (\textit{i.e.}, dichotomous image segmentation), (e) camouflaged object segmentation, (f) shadow detection, (g) concealed industrial defect detection, and (h) medical polyp lesion segmentation. The $\mathcal{M}$ represents mean absolute error (the lower the better). $\Delta$ shows the performance gaps between SAMs and the best performing state-of-the-art models.}
\vspace{-.1cm}
    \label{Tab1}
\end{table*}

\subsection{Numerical Results}
We report the quantitative results in Table~\ref{Tab1}.
To obtain SAM's results, following~\cite{tang2023can}, we first implement SAM\footnote{{\url{https://github.com/facebookresearch/segment-anything}}} to infer $N$ potential object masks within an input image. Given these masks, we choose the most suitable mask based on its alignment with the ground truth. Specifically, for $N$ binary predictions $\mathrm{\{P_n\}_{n=1}^{N}}$, and the ground-truth $\mathrm{G}$ for an input image, we compute intersection over union (IoU) scores for each prediction and ground-truth pair, generating a set of candidate scores ${\{IoU_n\}}_{n=1}^{N}$. The mask with the highest IoU is ultimately selected. The numerical findings presented in Table~\ref{Tab1} demonstrate a considerable performance gap between SAM and the top-performing model across eight benchmarks, especially for challenging scenarios and applications.
For instance, in the industrial application of Table~\ref{Tab1} (g), SAM with ViT-H attains a 0.265 MAE score, which is 17.6\% worse than the state-of-the-art DGNet~\cite{DGNet}. These results suggest that there is significant room for further research and enhancement of SAM's performance.

\section{Discussion and Outlook}
Here we discuss several potential directions for future research, as follows. 
\\
i) \textit{Application-oriented SAM \& dedicated large-scale dataset}: Although SAM achieves superior performance in natural image segmentation, it still delivers unappealing results in other applications, \eg, healthcare~\cite{MRNet21}, manufacture~\cite{he2019fully} and remote sensing~\cite{xu2018road}. The dedicated large-scale dataset and foundational model can be further explored to improve scalability in specific application scenarios. 
\\
ii) \textit{Additional prompt modes}: Current SAM primarily operates with three modes, click, box and everything. Some interesting prompt~\cite{xiang2022multimodal} can be further advanced by exploring, \textit{e.g.}, voice and gesture.
\\
iii) \textit{Pretraining strategy}: SAM's impressive success can be partially attributed to the massive segmentation dataset, SA-1B, which is better suited for pixel-wise prediction tasks compared to ImageNet. As such, SAM could serve as a novel pretraining model, with SA-1B functioning as a comprehensive pretraining dataset for industrial/medical applications. For instance, employing effective metric learning~\cite{kaya2019deep} on SA-1B could bolster the representation capabilities of features, allowing them to adapt to diverse tasks.
\\
iv) \textit{Multi-modal SAM}: Current SAM is less powerful for challenging scenarios such as low-contrast scenes and low-light nighttime cases. It is thus necessary to introduce complementary sources (\eg, depth~\cite{ji2022promoting,piao2019depth} and thermal~\cite{tu2019rgb}), which can effectively enhance the robustness of models.
\\
v) \textit{Video SAM}: Compared to static images, it is natural to consider dynamic video inputs~\cite{zhang2021dynamic,li2022exploring}. Therefore, how to develop a dynamic video SAM given initial prompt in first frame is an open issue.
\\
vi) \textit{Semi-supervised application}: The SAM has demonstrated exceptional performance and versatility,  making it a promising tool for various related tasks. Researchers can further leverage SAM to empower semi-supervised segmentation tasks~\cite{li2021joint}, using the model in combination with suitable point prompts, bounding box prompts, or scribble prompts to generate pseudo-labels.

\section{Conclusion}

In this paper, we conducted a preliminary investigation of SAM's performance across various applications, including natural image, agriculture, manufacturing, remote sensing, and healthcare. We analyzed SAM's benefits and limitations, identified potential challenges, and suggested future directions. These observations can guide the development of robust SAM algorithms and benchmarks. Moving forward, we plan to study SAM more in-depth, and design effective learning strategies to adapt it to a broader range of scenarios.
\\
\small{
\textbf{Acknowledgements.}
Thank Meta AI Research for the valuable and impressive work on providing open-source SAM model and SA-1B dataset. This study is partially supported by the Mitacs, CFI-JELF and NSERC Discovery grants. The views and conclusions contained in this paper are those of the authors and should not be interpreted as representing any funding agency.}

{\small
\bibliographystyle{ieee_fullname}
\bibliography{egbib}
}

\end{document}